\documentclass[10pt,twocolumn,letterpaper]{article}

\usepackage[pagenumbers]{cvpr} 

\usepackage[dvipsnames]{xcolor}

\usepackage{graphicx}
\usepackage{subcaption}
\usepackage{bm}
\usepackage{booktabs}
\usepackage{siunitx}
\usepackage{colortbl, color}
\usepackage{multirow}
\usepackage{tabularx}

\definecolor{cvprblue}{rgb}{0.21,0.49,0.74}
\usepackage[pagebackref,breaklinks,colorlinks,citecolor=cvprblue]{hyperref}

\def\paperID{*****} 
\def\confName{CVPR}
\def\confYear{2024}

\title{Event-Based Eye Tracking. AIS 2024 Challenge Survey}

\author{
Zuowen Wang$^{*}$ \and
Chang Gao$^{*}$ \and
Zongwei Wu$^{*}$ \and
Marcos V. Conde$^{*}$ \and
Radu Timofte$^{*}$ \and
Shih-Chii Liu$^{*}$ \and
Qinyu Chen$^{* \diamond}$ \and
Zheng-jun Zha\and
Wei Zhai\and
Han Han\and
Bohao Liao\and
Yuliang Wu\and
Zengyu Wan\and
Zhong Wang\and
Yang Cao\and
Ganchao Tan\and
Jinze Chen\and
Yan Ru Pei\and
Sasskia Brüers\and
Sébastien Crouzet\and
Douglas McLelland\and
Oliver Coenen\and
Baoheng Zhang\and
Yizhao Gao\and
Jingyuan Li\and
Hayden Kwok-Hay So\and
Philippe Bich\and
Chiara Boretti\and
Luciano Prono\and
Mircea Lic\u{a}\and
David Dinucu-Jianu\and
Cătălin Grîu\and
Xiaopeng Lin\and
Hongwei Ren\and
Bojun Cheng\and
Xinan Zhang\and
Valentin Vial\and
Anthony Yezzi\and
James Tsai
}

\begin{document}

\maketitle

\let\thefootnote\relax\footnotetext{$^*$ Zuowen Wang, Chang Gao, Zongwei Wu, Marcos V. Conde, Radu Timofte, Shih-Chii Liu and Qinyu Chen are the challenge organizers, while the other authors participated in the challenge. \\
$^{\diamond}$ Qinyu Chen (q.chen@liacs.leidenuniv.nl) is the corresponding author.\\
Challenge website: \url{https://eetchallenge.github.io/EET.github.io/}. Demonstration code repository: \url{https://github.com/EETChallenge/challenge_demo_code}. Challenge Kaggle website: \url{https://www.kaggle.com/competitions/event-based-eye-tracking-ais2024}. AIS 2024 host website: \url{https://ai4streaming-workshop.github.io/}.}

\begin{abstract}
This survey reviews the AIS 2024 Event-Based Eye Tracking (EET) Challenge. The task of the challenge focuses on processing eye movement recorded with event cameras and predicting the pupil center of the eye. The challenge emphasizes efficient eye tracking with event cameras to achieve good task accuracy and efficiency trade-off. During the challenge period, 38 participants registered for the Kaggle competition, and 8 teams submitted a challenge factsheet. The novel and diverse methods from the submitted factsheets are reviewed and analyzed in this survey to advance future event-based eye tracking research.  
\end{abstract}

\setlength{\abovedisplayskip}{1pt}
\setlength{\belowdisplayskip}{1pt}

\section{\textcolor{black}{Introduction}}

The fast development of augmented reality (AR) and virtual reality (VR) technologies in industry, 
has significantly magnified the importance of precise and efficient eye-tracking systems~\cite{fernandes2023leveling, fuhl2021teyed}. Furthermore, eye-tracking and related tasks, including gaze detection, pupil shape detection, etc, have tremendous potential in the field of wearable healthcare technology, offering novel approaches for diagnosing and monitoring conditions such as Parkinson's and Alzheimer's diseases through the analysis of eye movement patterns~\cite{pretegiani2017eye, duan2019dataset, Lee2000Schizophrenia}.

Energy consumption and computation resources are often constrained when implementing software and algorithms on mobile platforms. Moreover, the high sensory sampling rate for eye-tracking tasks also poses challenges from the sensor side. For mobile applications, ideally, the eye tracking system should be lightweight enough to fit in the headset while providing a high sampling rate.

Event cameras, or Dynamic Vision Sensors (DVS)~\cite{dvs1, dvs2, dvs3,zhou2023rgb}, provide a unique type of sensory modality for potential eye-tracking applications on mobile devices. Unlike traditional cameras that capture the entire scene synchronously at a fixed frequency, event cameras asynchronously record log intensity changes in brightness that exceed a threshold. This different sensing mechanism results in an inherently sparse spatiotemporal stream of output events, which, if appropriately exploited with the underlying algorithm~\cite{dvs_transformer2022Wang, Messikommer20eccv, Sabater_2022_CVPR, pointnet_event_camera,zhou2023dsec,zhou2023event,wu2024single} and computing platform~\cite{Aimar2017NullHopAF,chen2022skydiver,chen2020efficient,gao2022spartus}, can significantly reduce the computation and energy demands of the hardware platform. The high temporal resolution of the DVS events can also be useful for the eye-tracking task. 

This challenge aims to invite participants to explore algorithms for the event-based eye tracking task on a recorded eye tracking DVS dataset. By focusing on efficient algorithms capable of extracting meaningful information from sparse event streams, this challenge aims to pursue advancements in eye tracking technologies that are both energy-efficient and suitable for real-time applications in AR/VR technologies and wearable healthcare devices.

\section{\textcolor{black}{Event-based Eye Tracking Challenge}}

\subsection{\textcolor{black}{Introduction of the \text{3ET+} dataset}}
The \texttt{3ET+} dataset~\cite{wang2024ais_event} is an event-based eye-tracking dataset that contains real events recorded with a DVXplorer Mini~\cite{xplorer} event camera. There are 13 subjects in total, each having 2-6 recording sessions. The subjects are required to perform 5 classes of activities: \textit{random}, \textit{saccades}, \textit{read text}, \textit{smooth pursuit} and \textit{blinks}. The total data volume is 9.2 GB. The ground truth is labeled at 100Hz and consists of two parts for each label: (1) a binary value indicating whether there was an eye blink or not; (2) human-labeled pupil center coordinates.

\subsection{\textcolor{black}{Task description}} \label{sec:main_task_description}
\begin{itemize}
    \item \textbf{Input:} Raw event stream $(x_i, y_i, t_i, p_i)$ of recorded eye movement.
    \item \textbf{Task:}  Predict the pupil center spatial coordinate $(x,y)$ at required timestamps (same frequency as the ground truth) in the input space. 
    \item \textbf{Metric:}
The metric used in this work was also used in the work~\cite{3et}. The primary metric used on the Kaggle leaderboard was the p-accuracy. If the Euclidean distance between the spatial coordinates of the predicted label and the ground truth label is within p pixels, then we classify it as a successful prediction. On the Kaggle leaderboard, we set the tolerance as 10 pixels. We also provide p = \{5, 3, 1\} in the demonstration code pipeline for evaluating the model under a stricter tolerance value. After the Kaggle competition ended, we also provided a script for the participants to evaluate their results against the test set ground truth with mean Euclidean distance ($\ell_2$) and mean Manhattan distance ($\ell_1$). 
\end{itemize}

\subsection{\textcolor{black}{Provided pipeline for loading data and training}}
A handy data loading and training pipeline was provided to the challenge participants. The data loading module (data loader) is compatible with the Tonic library~\cite{tonic_lenz_gregor}. The module enables the participants to explore various event feature representations. 
The data loader can also cache the generated feature representation on the main memory or on disk during the first training epoch. For the following epochs, it can automatically load these pre-made data thus speeding up the training process. 

For the training pipeline, the challenge participants can easily place in their deep learning architecture and configure the hyperparameters. A machine learning monitoring library, namely the MLFlow library~\cite{mlflow}, was provided in the challenge pipeline code for the participants to monitor various metrics and to record the hyperparameters, as well as the checkpoints.

\subsection{Challenge phases}
The challenge is mainly divided into three phases: (1) before 5. Feb. 2024, preparation of challenge dataset, code pipeline, website, and Kaggle setup. (2) 5. Feb. 2024, the Kaggle competition begins, and teams are allowed to register and download the dataset. (3) 16. March. 2024, the submission system closed and the private score was released. The top-performing teams are then invited to submit their factsheets and every team was encouraged to submit a workshop challenge paper.

\section{\textcolor{black}{Challenge Results}}
The evaluation results of the final submissions from participating teams are listed in Tab.~\ref{tab:main_results}. The primary evaluation metric used in the Kaggle competition~\cite{eet_kaggle} ranking is the p10 accuracy described in Sec.~\ref{sec:main_task_description}. The challenge ranking is based on the private test split (p10 private (primary)), and only teams who submit their factsheets will participate in the final ranking. We also evaluate p10, p5, p3 and p1 accuracy and mean Euclidean distance ($\ell_2$) and mean Manhattan distance ($\ell_1$) based on the final submissions of the teams on the entire test split. Notably, all teams achieved very high p10 accuracy on the task.

\begin{table*}[h]
    \centering
    \footnotesize
    \begin{tabularx}{0.85\textwidth}{X | c | c |cccccc}
    \toprule
        \textbf{Team} & Rank & \textbf{p10 private (primary)}&p10 & p5 & p3 & p1 &$\ell_2$ & $\ell_1$ \\
        
     \midrule
     USTCEventGroup &  1 &99.58& 99.42& 97.05& 90.73& 33.75& 1.67& 2.11\\
     FreeEvs   & 2 &99.27& 99.26& 96.31& 83.83& 23.91& 2.03& 2.56\\
     bigBrains & 3 &99.16& 99.00& 97.79& 94.58& 45.50& 1.44& 1.82\\
     Go Sparse & 4 &98.74& 99.00& 77.20& 47.97& 7.32& 3.51& 4.63\\
     MeMo & 4 &98.74&99.05 & 89.36& 50.87& 6.53& 3.2& 4.04\\
     ERVT & 6 &97.60&98.21 & 94.94& 87.26& 28.80&1.98 &2.48 \\
     EFFICIENT & 6 &97.60& 97.95& 80.67& 49.08& 7.79& 3.51& 4.43\\
     GTechVision & 8 &91.86& 92.26& 61.08& 31.70& 4.16&4.94 & 6.18\\

     \bottomrule
    \end{tabularx}
    \caption{Final results from the top performing teams on Kaggle leaderboard. p10 private (primary) was evaluated on the private test set and all the rest metrics were evaluated on the entire test set.}
    \label{tab:main_results}
    \vspace{-4.mm}
\end{table*}
 
\subsection{Architectures and main ideas}
The participating teams proposed many different methods. This shows that there is no yet conclusive best approach for processing event data specifically for the eye tracking task. We summarize and discuss the major findings from the submitted methods.

\textbf{Stateful models and spatial-temporal processing.} Most teams selected architectures with state as their backbones, including gated recurrent units (GRU)~\cite{gru2014}, recurrent visual transformer~\cite{gehrig2023recurrent} with convolutional long short-term memory~\cite{shi2015convolutional} (ConvLSTM), bidirectional LSTM (biLSTM) and the newly emerged state-space model variant Mamba~\cite{gu2023mamba}. Stateful models are chosen due to the need to integrate event history and preserve the eye tracking states when very few events are generated, i.e., static or very slow eye movement. The other class of choices for handling the property of event data is using certain spatiotemporal techniques, including having a memory channel in the event representation preprocessing or temporal processing blocks in the convolution architecture.

\textbf{Computation and parameter efficiency.}
This challenge emphasizes efficiency and task performance as equally important. Most teams have considered computation and parameter efficiency in their method designs. One team (Sec.~\ref{sec:team:bigBrains}) designed a temporal causal layer that could be implemented with a buffer for online inference. The team Go Sparse (Sec.~\ref{sec:team:go_sparse}) implemented their method, which implements sparse convolution~\cite{choy20194d, choy2019fully}, on an FPGA board, achieving sub-millisecond inference latency. Most teams provided inference latency, model parameters and FLOPs needed for inference as part of their result.

\textbf{Event representations and spatial feature extraction.}
Converting raw event data into representations that can be processed by synchronous deep learning architectures is an important first step for any event-based tasks. In this challenge, the participants selected different ways to implement this step. Two teams aggregate sequential event frames using binary map representation to compress the input data sent to the model. The team MeMo (Sec.~\ref{sec:team:memo}) proposes using a memory channel to preserve historical event information better. Different from other teams, the team EFFICIENT (Sec.~\ref{sec:team:efficient}) uses point-based network~\cite{ren2024simple, qi2016pointnet, qi2017pointnet++} to process the raw event data as spatiotemporal event cloud. They also propose to use techniques that subsample the event cloud to reduce the computation. 

\textbf{Other novel components.} Apart from backbone design, hardware consideration, and event representation, the teams also propose other innovations during the challenge. This includes implementation of an affine transformation on event data (Sec.~\ref{sec:team:bigBrains}), sequence splitting and reordering data augmentation (Sec.~\ref{sec:team:memo}) and different regularization techniques (Sec.~\ref{sec:team:freeevs}, Sec.~\ref{sec:team:bigBrains}).

\subsection{Participants}
There were, in total, 38 user accounts registered and participated in the Kaggle competition~\cite{eet_kaggle} of the challenge, and 8 teams with private p10 accuracy over 90\% submitted factsheets describing their methods. The institutions of the team members for the submitted factsheets are located in regions including Asia, Europe and North America. 

\subsection{Inclusiveness and fairness}
Several measures were implemented to maintain the inclusiveness and fairness of the challenge. First and most importantly, the dataset and task are preprocessed and configured to keep the hardware requirement for training models possibly low. Second, an out-of-box usable training and testing pipeline was provided so that even teams with little event-based data experience could easily start experimenting. Thirdly, code submission is required for the factsheets submission to ensure reproducible results.

\subsection{Conclusions}
We summarize the insights obtained during the challenge and from the challenge results as follows:
\begin{enumerate}
    \item The field of event-based visual processing, more specifically event-based eye tracking, is a newly emerged field. Unlike many other computer vision fields where transformer-based methods dominate, there is a large variety in event data processing. There is plenty of room for achieving a better trade-off for task performance and model efficiency.
    \item Hardware consideration is always essential for researchers developing algorithms for event cameras due to its compatibility with mobile platforms. Algorithm-hardware co-design is an important research direction in this field.
    \item This challenge and existing works proved the feasibility of using an event camera for the eye-tracking task. Prototyping and more realistic settings are needed to step towards more mature event-based eye tracking systems. 
    
\end{enumerate}

\paragraph{Related Challenges} This challenge is one of the AIS 2024 Workshop associated challenges on: Event-based Eye-Tracking~\cite{wang2024ais_event}, Video Quality Assessment of user-generated content~\cite{conde2024ais_vqa}, Real-time compressed image super-resolution~\cite{conde2024ais_sr}, Mobile Video SR, and Depth Upscaling.

\section*{Acknowledgements}

Zongwei Wu, Marcos V. Conde, and Radu Timofte are with University of W\"urzburg, CAIDAS, Computer Vision Lab.

Zuowen Wang and Shih-Chii Liu are at the Institute of Neuroinformatics, University of Zurich and ETH Zurich (INI).
Qinyu Chen was at INI at the start of the challenge and is currently at the Leiden Institute of Advanced Computer Science (LIACS), Leiden University.

Chang Gao is at Delft University of Technology.

This project was partially (e.g., dataset collection) funded by the Swiss National Science Foundation and Innosuisse BRIDGE - Proof of Concept Project (40B1-0\_213731).

The dataset collection was partially supported by the 2023 Telluride Neuromorphic Cognition Engineering Workshop.

This work was partially supported by the European Union's Horizon 2020 research and innovation programme
under grant agreement No 899287.

This work was partially supported by the Humboldt Foundation. We thank the AIS 2024 sponsors: Meta Reality Labs, Meta, Netflix, Sony Interactive Entertainment (FTG), and the University of W\"urzburg (Computer Vision Lab).

\section{Challenge Methods and Teams}
\label{sec:teams}

In the following sections we describe the best challenge solutions. Note that the method descriptions were provided by each team as their contribution to this survey. 

\subsection{Team: USTCEventGroup}


\begin{center}

\noindent\emph{Zheng-jun Zha, Wei Zhai, Han Han,\\Bohao Liao, Yuliang Wu}

\noindent\emph{University of Science and Technology of China}

\noindent{\emph{Contact: \url{zhazj@ustc.edu.cn}}}

\end{center}

\begin{figure}[t]
    \centering
\includegraphics[width=1.0\linewidth]{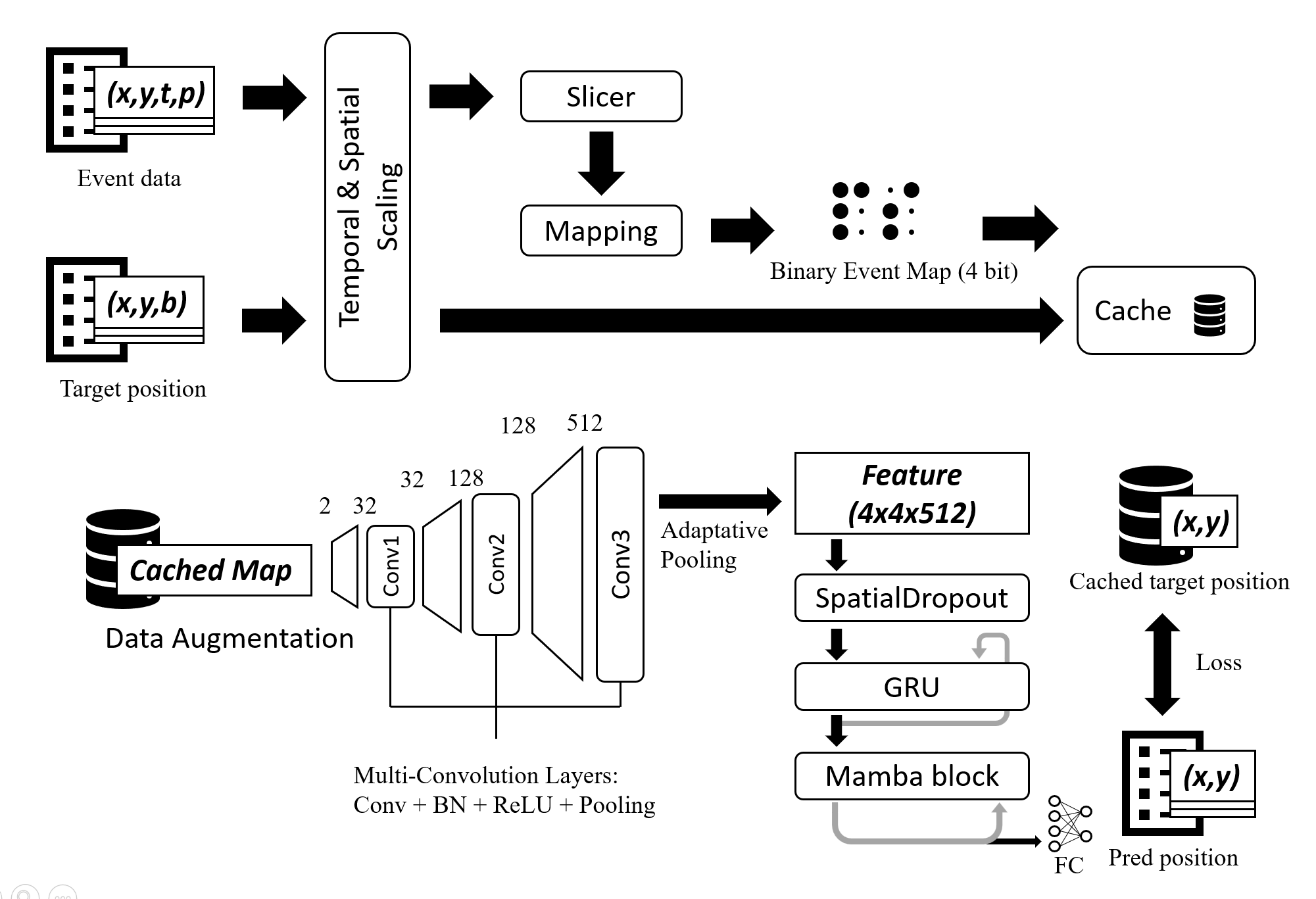}
    \caption{MambaPupil network by Team USTCEventGroup.}
    \label{fig:MambaPupil-network}
\end{figure}

\paragraph{Description.} 
The USTCEventGroup proposed the MambaPupil method as shown in \cref{fig:MambaPupil-network}. This lightweight and time-efficient network comprises two main parts: the Spatial Feature Extractor and the Dual-Recurrent Module for position prediction. In the Spatial Feature Extractor there are a series of convolutional blocks, which are designed as follows: 
\begin{align}
    x_t = Pool(ReLU(BatchNorm(Conv(B_e)))),
\end{align}
Larger convolutional kernels (7 or 5) are employed, and extracted features are fed into the Dual Recurrent Module after an adaptative global pooling layer and a SpatialDropout layer. The Dual Recurrent Module consists of a bi-directional GRU module and an LTV-SSM module, recently known as a Mamba block. This bi-directional structure is beneficial to capturing contextual information compared to a uni-directional one, and the additional Linear-time-variant State-Space Model (LTV-SSM) module models the behavior patterns of eye movements selectively to cast more attention into the valid phase. The LTV-SSM module can be described by the formula:
\begin{align}
    \Delta, B, C &= Linear(x), \\
    \Delta A &= exp(\Delta * A), \\
    \Delta B &= \Delta * B. \\
    \Delta x &= \Delta A * x + \Delta B *u, \\
    y &= Cx + Du,
\end{align}

Furthermore, to reduce computational complexity and avoid uncorrelated events negatively affecting the model, The USTCEventGroup utilizes binary map representation of events and encodes the order of events in the sequence. The binary map representation, usually referred to as bina-rep, is aggregated from multiple sequential event frames. These frames are firstly binarized and stacked by bits forming a bina-rep. Additionally, several data enhancement methods are introduced, including Event-Cutout, which applies spatial random masking to the event image and extends it across the entire sequence in the data augmentation stage to enhance the generalization ability. As for the loss criterion, RMSE loss is applied:

\begin{align}
    Loss = \sqrt{\frac{1}{L}\sum_{i=1}^{L}(Pred_{i} - Label_{i})^2} ,
\end{align}

\paragraph{Implementation Details.} All the experiments were implemented using PyTorch, and Cosine Annealing Warm Restart was utilized as a learning-rate scheduler, beginning at an initial rate of 0.002. The training and evaluation are performed entirely on the 3ET++ dataset, and the time cost for 1000 epochs with a batchsize of 32 is about 1.5 hours on a single RTX 2080Ti GPU.


\paragraph{Results.}

We scored best on the leaderboard, performed far exceeded the baseline with a p10 accuracy above 0.995. Our visualization shows the elegant tracking process even in difficult scenario. Visualization results and code is available in \href{https://github.com/hh-xiaohu/Event-based-Eye-Tracking-Challenge-Solution}{https://github.com/hh-xiaohu/Event-based-Eye-Tracking-Challenge-Solution}.

The results are obtained from the validation set:

\begin{table}[h]
\centering
    \begin{tabular}{c|c|c|c|c}
        \hline
        & p\_5 & p\_10 & p\_15 & p\_error\\
        \hline
       MambaPupil & 0.9192 & 0.9846 & 0.9926 & 2.1799 \\
       \hline
        
    \end{tabular}
    \caption{Validation results of MambaPupil.}
    \label{tab:ustceventgroup:valid_results}
\end{table}



\subsection{Team: FreeEvs}
\label{sec:team:freeevs}

\begin{center}

\noindent\emph{Zengyu Wan, Zhong Wang, Yang Cao,\\Ganchao Tan, Jinze Chen}

\noindent\emph{University of Science and Technology of China}

\noindent{\emph{Contact: \url{wanzengy@mail.ustc.edu.cn}}}

\end{center}

\begin{figure}[t]
    \centering
    \includegraphics[width=0.95\linewidth]{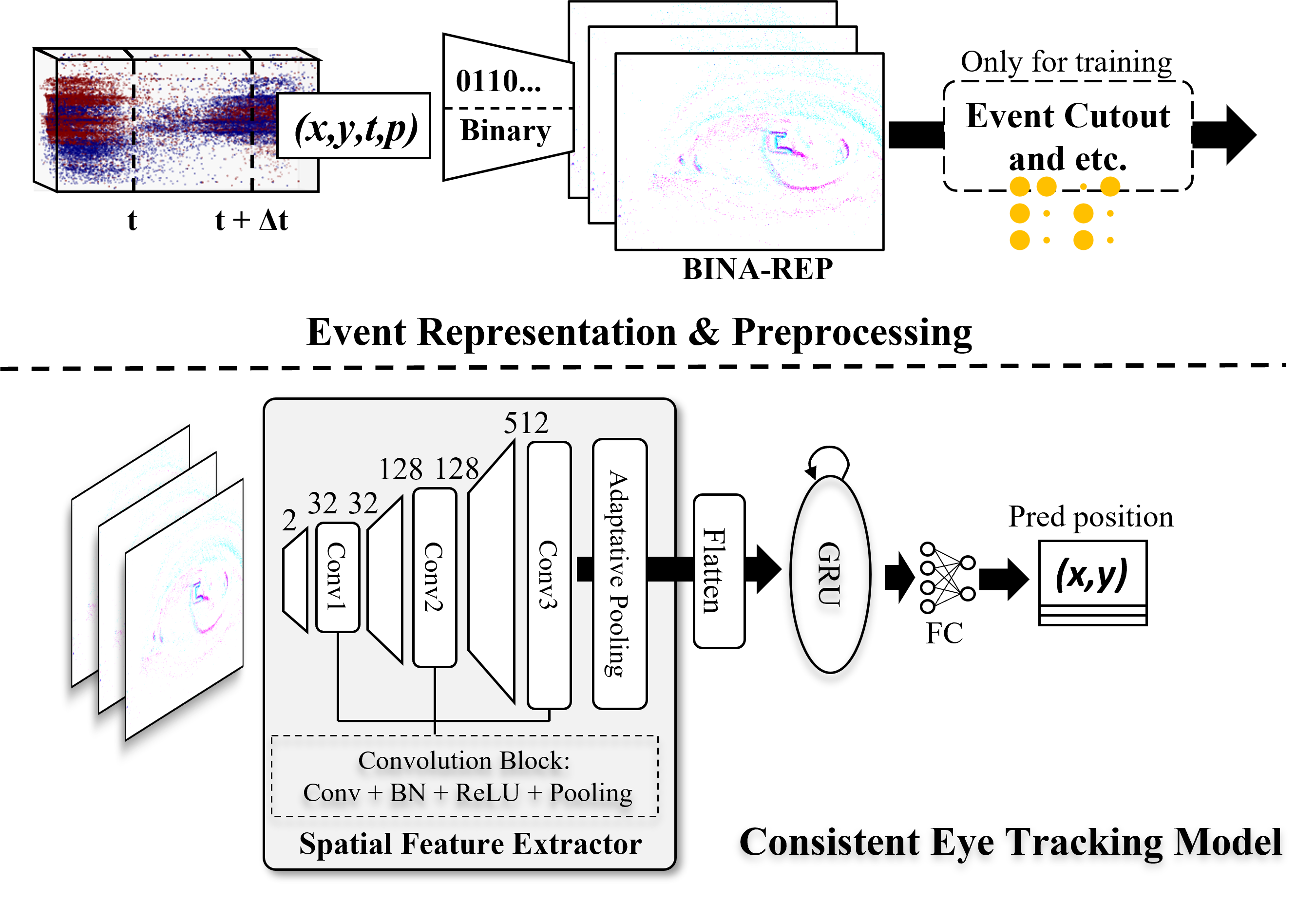}
    \caption{Schematic of the Consistent Eye Tracking Model, including the event representation, preprocessing and the tracking predictor.}
    \label{fig:CETM}
\end{figure}

\paragraph{Description.} The FreeEvs Team proposed the lightweight and efficient model, Consistent Eye Tracking Model (CETM), as shown in \ref{fig:CETM}. The whole framework consists of the following components: 1. Event representation; 2. Representation enhancement; 3. Tracking predictor; 4. Motion consistency loss. Given the event stream, the CETM will first convert it into the binary map representation \cite{bina-rep}, which is compact and informative. The binary map first generates a spatially binarized event clip based on the event frames aggregated and accumulated over a period of time, after which it encodes each frame in the clip using a temporal binarization mask. And then, the spatio-temporally binarized encoded clip is accumulated at corresponding locations to form the final Bina-rep $B_e$. This approach brings two benefits: 1) it reduces storage and computational costs, improving the training and inference speed of the network.; 2) it reduces the impact of noise and redundant information on prediction results to some extent. Next, the Bina-rep will be enhanced for effective training. The adopted data enhancement techniques contain spatial shift, spatial flip, temporal shift, and temporal flip. The spatial operations improve the model localization ability with more spatial variation samples, and the temporal operations endow the model's consistent tracking ability with more temporal variation samples. Next, the tracking predictor is utilized to localize the pupil efficiently. The predictor consists of the stacked convolution blocks, the global pooling layer, the gated recurrent unit (GRU), and the final full connection layer (FC) to predict the result. Among them, the team used 3 convolution blocks, each composed of the 2D convolution, batch normalization, ReLU activation function, and average pooling layer, to extract the corresponding pupil space information. And then, the spatial information will be transported into the GRU module in time order to extract the temporal contextual clues. The spatial-temporal clues are finally transformed into the prediction by the FC layer. In the training stage, the motion consistency loss $L_{mc}$ is adopted to supervise the tracking result not only on the original prediction (the localization) but also on the one-order difference result for smoothness. The motion consistency is formulated as follows,
\begin{align}
    &L_{mc} = L_0 + L_1, \\
    &L_0 = \sqrt{(x_{pred} - x_{gt}) ^ 2 + (y_{pred} - y_{gt}) ^ 2},\\
    &L_1 = \sqrt{(\Delta_t x_{pred} - \Delta_t x_{gt}) ^ 2 + (\Delta_t y_{pred} - \Delta_t y_{gt}) ^ 2}, \\
    &\Delta_t p = p_{t} - p_{t-1},
\end{align}
where the $x_{pred}, y_{pred}$ are the spatial prediction, $x_{gt}, y_{gt}$ are the corresponding ground truth. The motion consistency loss endows the model smoothing tracking ability by supervised constraining the change in prediction over time. 


\paragraph{Implementation Details.}
The whole framework is implemented by PyTorch and the experiments were conducted on a single NVIDIA RTX2080Ti GPU. During training, the Adam optimizer is adopted, and the learning rate is set to 5e-4 with a batch size of 8. And the training epoch is set to 800 for full training. For efficiency consideration, the resolution of the input event is downsampled to 80x60. In all, the proposed CETM has 7.1M parameters and needs 2.9G flops in the inference stage.



\paragraph{Results.}

Our method eventually achieved 99.27 accuracy on the test dataset, ranking third on the benchmark. The results in detail are below.

\begin{table}[h]
\centering
    \begin{tabular}{c|c|c|c|c}
        \hline
        &P5 acc & P10 acc & P15 acc & Euc Dist\\
        \hline
       CETM & 91.8 & 97.6 & 98.5 & 2.50 \\
       \hline
        
    \end{tabular}
    \caption{Validation results of Team FreeEvs.}
    \label{tab:FreeEvs:valid_results}
\end{table}

\subsection{Team: bigBrains} \label{sec:team:bigBrains}


\begin{center}

\noindent\emph{Yan Ru Pei, Sasskia Brüers, Sébastien Crouzet, Douglas McLelland, Olivier Coenen}

\noindent\emph{Brainchip Inc.}

\noindent{\emph{Contact: \url{ypei@brainchip.com}}}

\end{center}


\paragraph{Description.} 

\begin{figure*}[htbp]
  \centering
  \includegraphics[width=\linewidth]{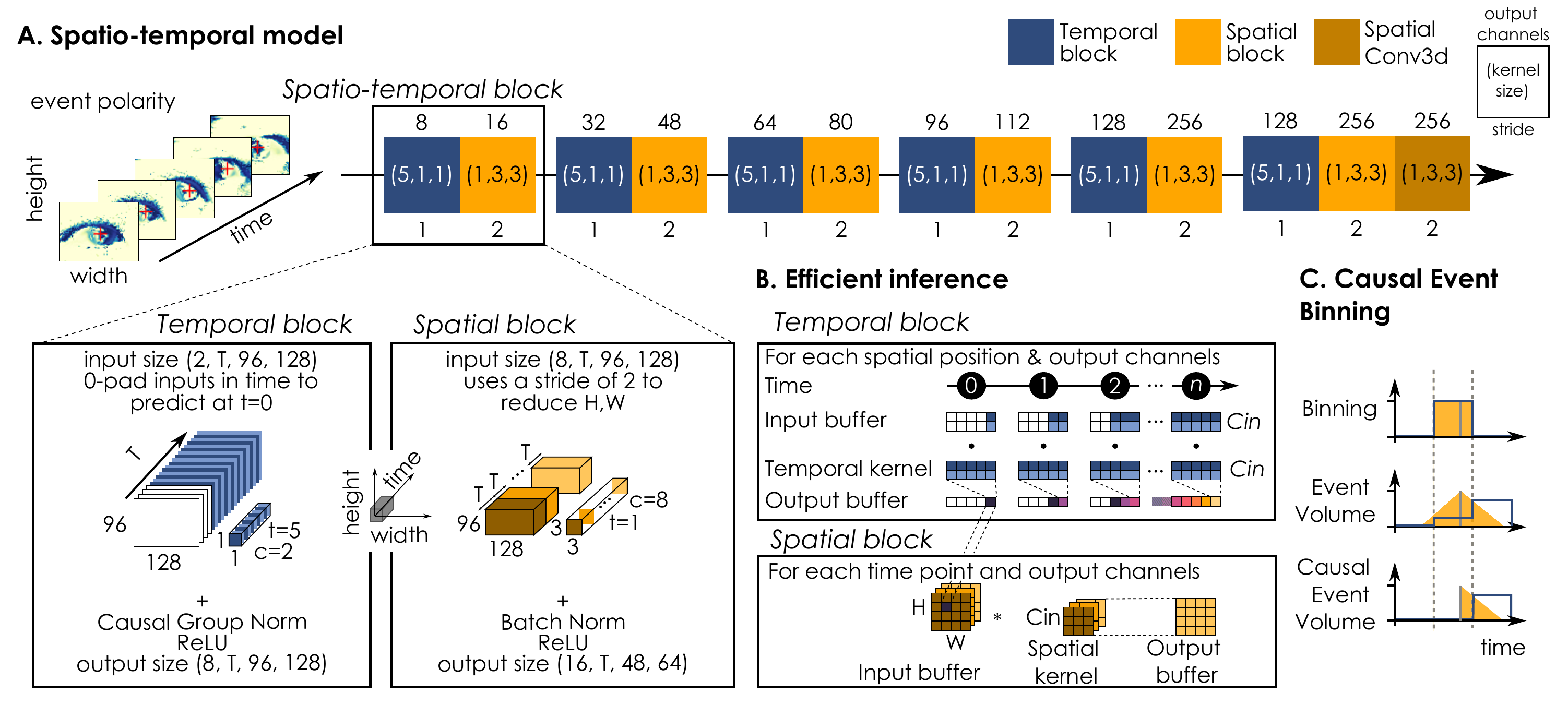}
  \caption{\textbf{A.} A lightweight spatiotemporal architecture for efficient eye tracking. The backbone is composed of a succession of 5 spatiotemporal blocks. Each spatiotemporal block consists of a temporal convolution followed by a spatial convolution. \textbf{B.} The model can be configured to run in streaming inference mode by using an input FIFO buffer for each temporal layer. The sliding-window mechanism of the FIFO buffer would act as the convolution sliding window, and the convolution operation itself is simply replaced by a dot product between the elements in the FIFO buffer and kernel weights. \textbf{C.} Compares the methods of direct binning, event volume binning, and causal event volume binning. The last method retains temporal information while still being fully causal.}
  \label{fig_st_network}
\end{figure*}

The bigBrains team from Brainchip Inc. proposed a lightweight spatio-temporal network for online eye tracking with event camera \cite{pei2024spatiotemporal}, depicted in Fig.~\ref{fig_st_network}.A. We use a causal spatio-temporal convolutional network (with (1+2)D-style factorized temporal plus spatial convolutions) that can be configured to perform online inference very efficiently on streaming data (i.e. event streams). The detector head of the network consists of two spatial convolutional layers inspired by CenterNet. Our temporal layers are causal, meaning that during inference, the layers can accept streaming input features and store them using a FIFO buffer (Fig.~\ref{fig_st_network}.B). For data processing, we convert the event data into 10 ms ``frames" using a causal event-volume binning strategy to minimize latency for online inference (Fig.~\ref{fig_st_network}.C). Data augmentation during training includes spatio-temporal affine augmentations directly on the events (including temporal + polarity flips, but excluding spatial reflections).

\paragraph{Implementation Details.}
For training, we use a batch size of 32, where each batch consists of 50 event frames. We train for 200 epochs with the AdamW optimizer with a base learning rate of 0.002 and a weight decay of 0.005. We use the cosine decay with linear warmup scheduler, with warmup steps equal to 0.025 of the total training steps. We use automatic mixed-16 precision along with PyTorch compilation during training.

Note that our model is trained with the temporal dimension (number of frames) $T=50$, but this number can be anything. For training, a longer $T$ will reduce the implicit pre-padding artifacts of a sample, but will reduce the robustness of the batch statistics (assuming batch size goes down with increasing $T$). For inference and model evaluation, we feed entire segments to the network without splitting because we want to eliminate the artifacts of implicit padding as our network is agnostic to the number of input frames. 

The training is done on a single NVIDIA A30 GPU, wihch took 0.309 hours for a total of 200 epochs. The streaming inference is done on a single thread of an AMD EPYC 7543P processor (CPU) which takes 3.384 ms per frame, with an event frame corresponding to a 10 ms time-window binning of events. The inference pipeline is shared with the training pipeline in PyTorch, and is still highly unoptimized. For inference on the NVIDIA A30 GPU, it takes 659 ms per 1000 frames.

Our backbone model used during development contains only 164 thousand parameters, and by itself was able to perform quite well on the test metrics (same position on the private leaderboard). For submission however, we chose to use a larger backbone combined with a heavily parameterized detection head, yielding a total of 1.1 million parameters, which is the largest model we have tested. In our report, we will provide a detailed ablation study testing the model variants.

We measured activation sparsity in the network (over the validation set), and found it to average around 50\%, about what one would expect given ReLU activation functions. Reasoning that much of this activity may not be informative, given the very high spatial sparsity of input to the network, we applied $L_1$ regularization to activation layers. By adding this regularization loss with a weighing factor of 0.1 to the total loss function, the network can achieve a sparsity of 90\% while suffering little performance drop.

The final output feature of the network is a $3 \times 4$ grid overlaid on top of the event frames, each grid-cell in each frame containing a prediction of: 1) the probability of a pupil being inside the cell, and 2) the relative x and 3) y offset of the pupil in the cell. We apply the following loss function to each grid-cell
\begin{equation*}
\text{loss} = 
\begin{cases} 
-(1 - \hat{p})^{\gamma} \log(\hat{p}) + \text{regression\_loss} & \text{if } p = 1 \\
\hat{p}^{\gamma} \log(1 - \hat{p}) & \text{if } p = 0
\end{cases},
\end{equation*}
where the focal loss parameter $\gamma=2$ and $\hat{p}$ is the predicted probability of pupil presence and $p$ is the ground truth presence. The regression loss is the summed SmoothL1Loss for the $\hat{x}$ and $\hat{y}$ offset predictions of the network in each grid cell. For example, the SmoothL1Loss between the $\hat{x}$ prediction and the $x$ ground truth is, 
\begin{equation*}
\text{SmoothL1Loss}
=
\begin{cases} 
0.5 \times \frac{(\hat{x} - x)^2}{\beta} & \text{if } |\hat{x} - x| < \beta \\
|\hat{x} - x| - 0.5 \times \beta & \text{otherwise}
\end{cases},
\end{equation*}
where $\beta = 0.11$. The total loss is computed by averaging over all grid-cells and valid frames (where the eye is open and within bounds).

\paragraph{Results.}

\begin{table}[htbp]
\centering
\caption{\textbf{Ablation study results for modifications of the model architecture and event processing methods.} The default parameters are highlighted in \textit{italics}. \textit{CenterNet with DWS temporal} means the temporal smoothing layer before the head is depthwise-separable. MACs is the number of multiply-accumulate operations done per event frame, not accounting for sparsity. Results are the average of 10 repeats.}
\label{table:ablation-model}
\begin{tabular}{@{}lllll@{}}
\toprule
 & \textbf{p10} & \textbf{dist.} & \textbf{params} & \textbf{MACs} \\
\midrule
\textit{Benchmark}           & 0.963 & 2.79 & 809K & 55.2M \\
                             &       &      \\
\multicolumn{5}{l}{\textbf{Event-processing} - \textit{Causal event volume}} \\
$\rightarrow$ Event volume   & 0.959 & 2.77  & - & - \\
$\rightarrow$ Binning        & 0.959 & 2.74  & - & - \\
                             &       &      \\      
\multicolumn{5}{l}{\textbf{Model head} - \textit{CenterNet with DWS temporal}} \\
$\rightarrow$ Full temporal  & 0.964 & 2.72 & 1.07M & 58.3M \\
$\rightarrow$ No detector    & 0.936 & 3.52 & 216K & 47.4M \\
                             &       &      \\
\multicolumn{5}{l}{\textbf{Temporal kernel size} - \textit{5}}   \\
$\rightarrow$ 3       & 0.955  & 3.20 & 801K & 46.9M \\
                      &  & \\
\multicolumn{5}{l}{\textbf{Spatiotemporal block} - \textit{(1+2)D}} \\
$\rightarrow$ Conv3D    & 0.969 & 2.50 & 1.21M & 267M \\

\bottomrule
\end{tabular}
\end{table}

Table~\ref{table:ablation-model} reports the validation metrics for various configurations. We see that the event processing choices had very little impact on accuracy, but our benchmark (with causal event volume binning) ensures causality (see Fig.~\ref{fig_st_network}.C). For network architectural choices, we see that using a CenterNet like head led to a boost of 0.027, compared with the ``no detector" (a simpler head with global average pooling and 2 dense layers). Similarly, using a larger temporal kernel led to a small boost in p10 accuracy (0.008) at a small computation cost (8.3M MACs). Using full 3D convolutions can boost performance slightly (Table~\ref{table:ablation-model}) but at the cost of significantly increased memory and computational load.

\begin{figure*}[t]
  \centering
  \includegraphics[scale=0.5]{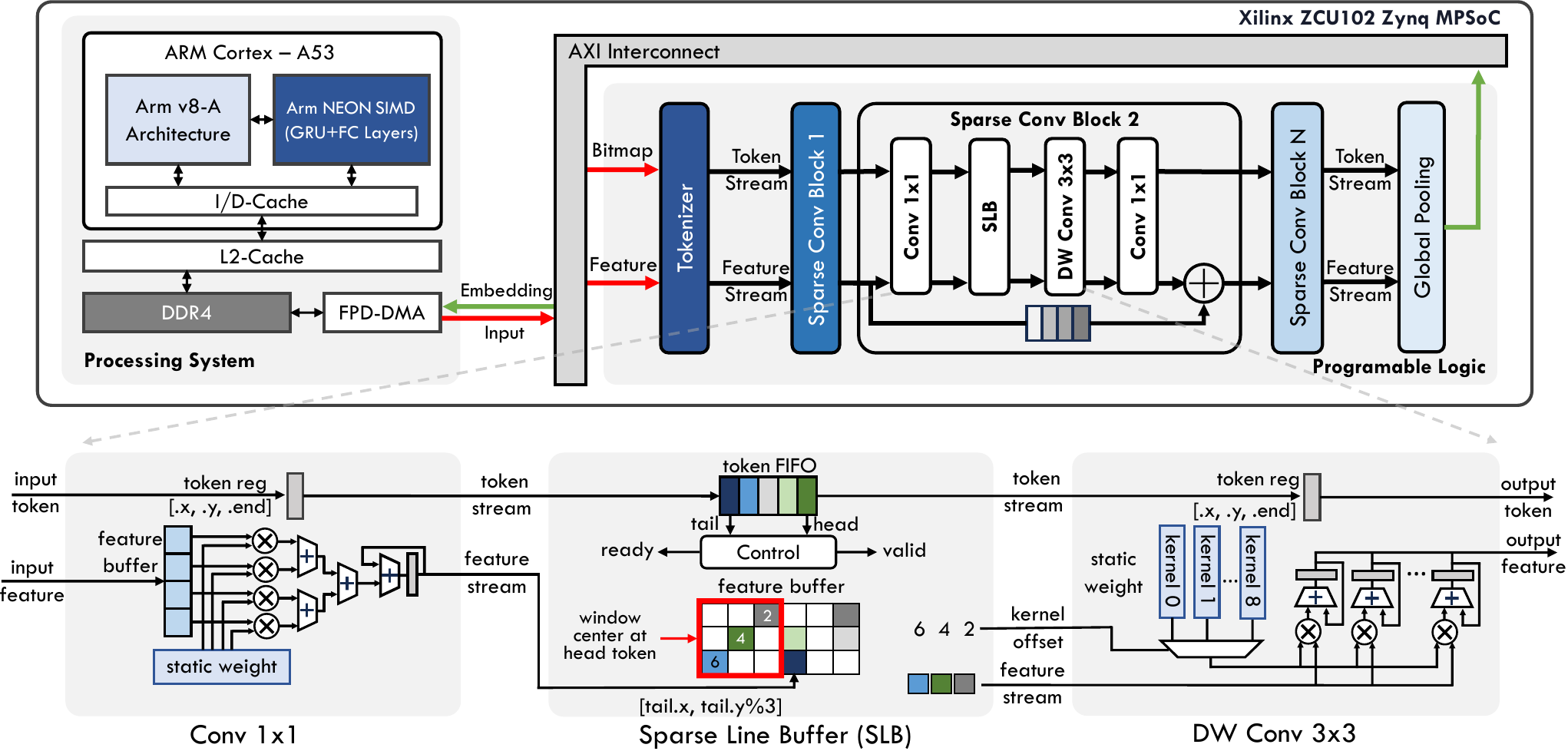}
  \caption{Hardware design of SEE}
  \label{fig:gosparse:heterogeneous}
\end{figure*}

\subsection{Team: Go Sparse} \label{sec:team:go_sparse}


\begin{center}
\noindent\emph{Baoheng Zhang, Yizhao Gao, Jingyuan Li,\\ Hayden Kwok-Hay So}

\noindent\emph{The University of Hong Kong}

\noindent{\emph{Contact: \url{bhzhang@eee.hku.hk}}}   
\end{center}

\newcommand{\aname}{SEE\xspace}

\paragraph{Description.} 
We propose an eye-tracking system deploying on FPGA achieving 1ms latency. The model uses a lightweight MobilenetV2-based architecture to extract features, which will then be fed to a GRU and a fully connected layer to generate the eye center location. We conduct data augmentation like random shifting and flipping to boost the performance. The system is co-designed with submanifold sparse convolution that only processes non-zero spatial feature activations. We conduct integer-only quantization allowing the overall design to be more efficient. The hardware dataflow accelerator can leverage the sparsity and deliver low-latency, low-power performances, achieving sub-millisecond inference for each prediction.

\paragraph{Implementation Details.}
To address the eye-tracking problem efficiently, we propose SEE, a hardware-software co-optimization solution. On the software side, our model comprises a SCNN-based backbone for feature extraction, a GRU layer for temporal feature fusion, and a fully connected (FC) layer for eye center regression. Our hardware is heterogeneous, as the FPGA programmable fabric is used for SCNN acceleration and Arm Cortex-A series for GRU and FC layers. This heterogeneous architecture allows us to fully exploit the strengths of different hardware devices and deliver an overall low-latency performance. In addition,  we also employ hardware-software co-optimization to search for compact models with better tradeoffs between accuracy and hardware latency.

\begin{figure}
  \centering
  \includegraphics[scale=0.12]{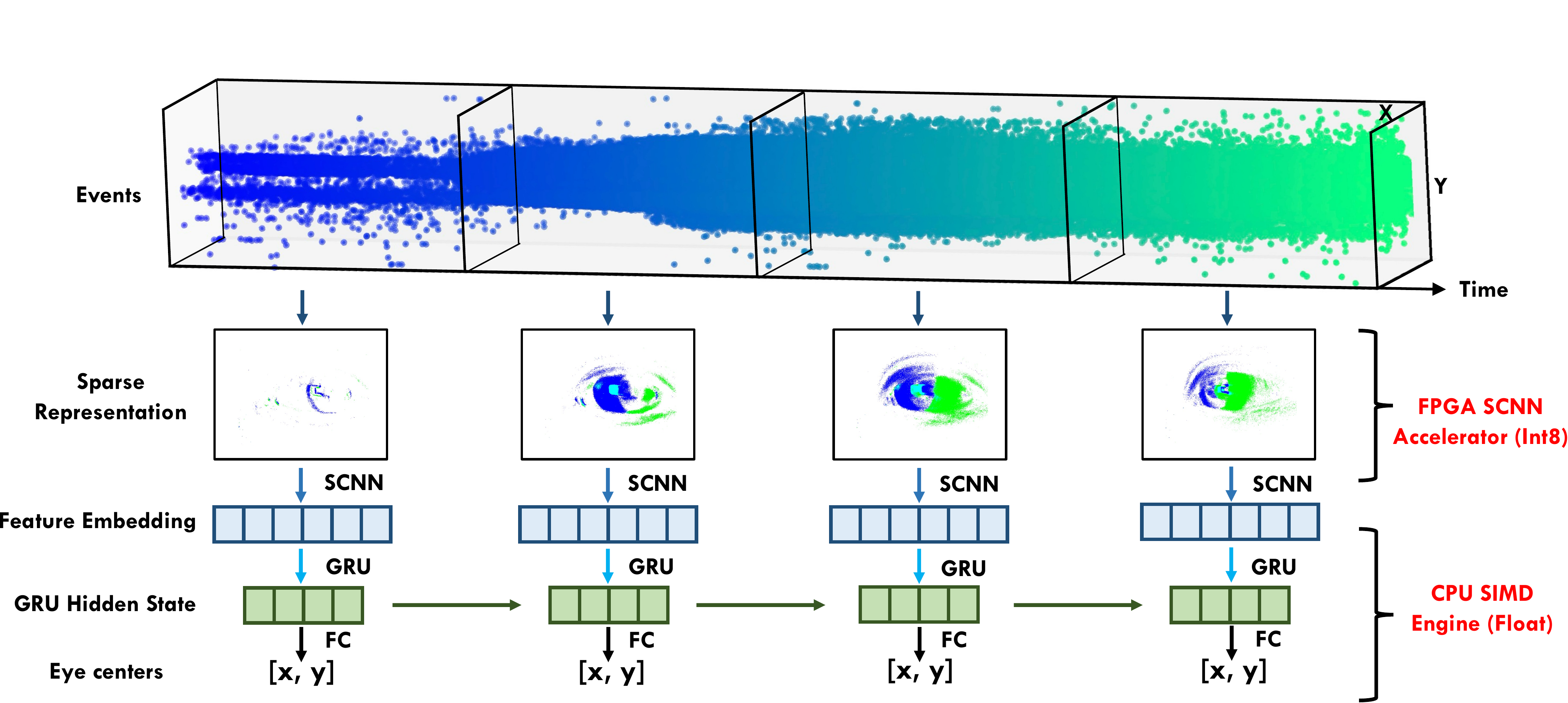}
  \caption{Software architecture of SEE}
  \label{fig:gosparse:sw_arch_crop}
\end{figure}

The software architecture is depicted in Figure \ref{fig:gosparse:sw_arch_crop}, the event clips in a fixed-time interval usually are spatially sparse, which means most of the pixels are completed zero. These sparse inputs are fed into the SCNN backbone to extract global features.
Subsequently, these features undergo further processing through a GRU layer, which captures the temporal information between event frames. The hidden features are then fed into the FC layer, yielding the normalized coordinates of the eye center location, ranging from 0 to 1. The actual eye location pixel coordinates can be obtained directly by multiplying these normalized coordinates with the height and width of the input size.

\begin{table}[]
\centering
\footnotesize
\caption{Implementation Details of SEE.}
\begin{tabular}{cccccc}
\hline
 &
  \multicolumn{2}{c}{Accuracy (\%)} &
   &
   &
   \\ \cline{2-3}
\multirow{-2}{*}{} &
  p5 &
  p10 &
  \multirow{-2}{*}{\begin{tabular}[c]{@{}c@{}}Dist.\\ (Pixel)\end{tabular}} &
  \multirow{-2}{*}{\begin{tabular}[c]{@{}c@{}}\#\\ Param.\end{tabular}} &
  \multirow{-2}{*}{\begin{tabular}[c]{@{}c@{}}Latency\\ (ms)\end{tabular}} \\ \hline
MobileNetV2 &
  {\color[HTML]{000000} \textbf{87.36}} &
  {\color[HTML]{000000} 99.53} &
  {\color[HTML]{000000} \textbf{3.15}} &
  797K &
  1.45 \\ \hline
SEE-A &
  {\color[HTML]{000000} 80.83} &
  {\color[HTML]{000000} \textbf{99.60}} &
  {\color[HTML]{000000} 3.77} &
  465K &
  {\color[HTML]{000000} 0.64} \\
SEE-B &
  {\color[HTML]{000000} 83.32} &
  {\color[HTML]{000000} 99.53} &
  {\color[HTML]{000000} 3.39} &
  372K &
  {\color[HTML]{000000} 0.94} \\
SEE-C &
  \cellcolor[HTML]{FFFFFF}{\color[HTML]{000000} 75.92} &
  \cellcolor[HTML]{FFFFFF}{\color[HTML]{000000} 98.39} &
  \cellcolor[HTML]{FFFFFF}{\color[HTML]{000000} 4.05} &
  180K &
  {\color[HTML]{000000} \textbf{0.60}} \\
SEE-D &
  {\color[HTML]{000000} 81.37} &
  {\color[HTML]{000000} 99.53} &
  {\color[HTML]{000000} 3.71} &
  \textbf{178K} &
  {\color[HTML]{000000} 0.70} \\ \hline
\end{tabular}
\label{tab:hardware_performance}
\end{table}

The hardware implementation is shown in \ref{fig:gosparse:heterogeneous} built upon a Xilinx Zynq UltraScale+ MPSoC device. The proposed hardware system primarily consists of two components: the sparse dataflow SCNN accelerator and the Arm Cortex-A53 processor host. The event-based input is initially fed into the SCNN accelerator to propagate through the submanifold sparse convolutional~\cite{choy20194d, choy2019fully} neural network backbone. Subsequently, the GRU and fully connected layers processes are executed by the host CPU with the Arm NEON SIMD (Single Instruction, Multiple Data) engine. 


\paragraph{Results.}
We trained MobileNetV2 (width multiplier = 0.5) and our \aname-series models with 4 different structures. The measured accuracy and efficiency are shown in Tab. \ref{tab:hardware_performance}.
When evaluating with  p10 accuracy, we observe that our MobileNetV2 and the \aname-series networks achieve comparable high accuracies, mostly exceeding 98\%. While considering the p5 accuracy and the mean Euclidean distance, the baseline MobileNetV2 slightly outperforms the \aname-series models. This difference could be attributed to the higher number of network parameters since a larger model size generally provides more capacity to capture richer features.

In terms of efficiency, our selected \aname-series model significantly outperforms MobileNetV2 by a large margin. MobileNetV2 achieves a latency of 1.4 ms, which is more efficient than the previous work. However, our \aname-series model can even achieve a latency of less than 1 ms. Specifically, our \aname-D model achieves a comparable accuracy with MobileNetV2 with $2\times$ speedup (0.7 ms vs. 1.45 ms). Our \aname-C model (0.6 ms) achieves around $2.5\times$ speedup over MobileNetV2 with only 1\% p10 accuracy drops. This highlights the capability of our \aname-framework to push more optimal latency accuracy trade-offs than baseline. 

\subsection{Team: MeMo}
\label{sec:team:memo}


\begin{center}
\noindent\emph{Philippe Bich, Chiara Boretti, Luciano Prono}

\noindent\emph{Polytechnic of Turin}

\noindent{\emph{Contact: \url{philippe.bich@polito.it}}} 
\end{center}

\begin{figure}[t]
    \centering
    \includegraphics[width=3.5in]{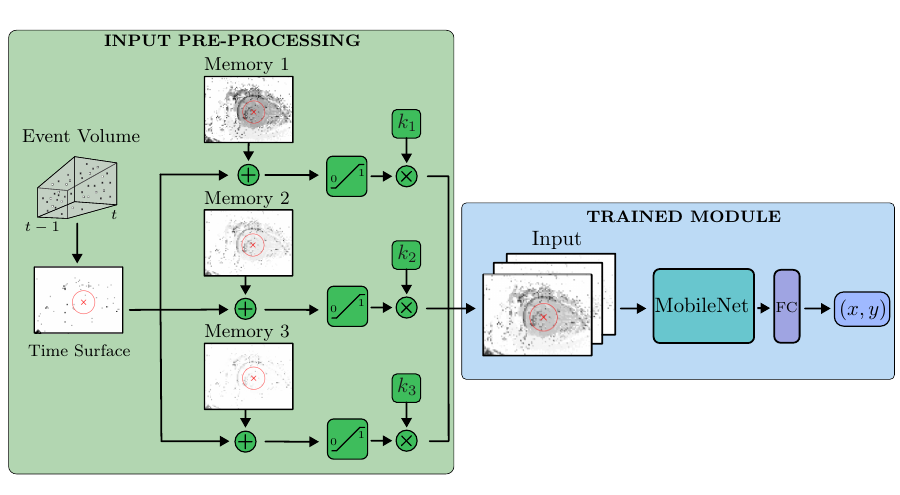}
    \caption{Schematic of the method used to solve the event-based eye-tracking task. Starting from a volume of events collected in the time range $(t-1, t)$, a time surface is created and then enriched with memory channels built starting from events that were collected in the time range $(0, t-1)$. The input of the network is then the concatenation of the three enriched time surfaces while the output is the pupil's position where $x, y \in [0,1]$. In this work $k_1=0.8$, $k_2=0.6$ and $k_3=0.4$. The final FC block in the submitted solution includes two fully-connected layers with 220 and 2 neurons respectively.}
    \label{fig:trashcoders:base}
\end{figure}

\paragraph{Description.} 
We present an eye-tracking system depicted in Figure~\ref{fig:trashcoders:base}, composed of two main parts: an input pre-processing pipeline and a model for estimating the eye's pupil position. 
Our methodology centers on the concept of memories of events, wherein we integrate the events in multiple surfaces to obtain enriched input data. Starting from the events stream, we isolate the events in a time range $\Delta t$ set to $50ms$ (20Hz) and we represent positive and negative events using two independent time surfaces \cite{7508476} that are then averaged obtaining a tensor of shape $[1, H, W]$ with $H=60$ and $W=80$.

Unfortunately, not every time surface obtained like this 
contain enough information to estimate the position of the eye's pupil in the given time range. One possible solution to this problem is the usage of recurrent neural networks (RNNs) \cite{3et}. However, RNNs are typically more complex and computationally intensive to train compared to non-recurrent models. Therefore, our proposed solution enables the use of non-recurrent DNNs for estimating the position of the eye's pupil.
This can be done by enriching the information of the events collected in the time range $\Delta t$ with older events that are stored in multiple memories with different forgetting rates. Following this idea, the input of the DNN consists of multiple channels generated with a negligible time overhead as
        
\begin{equation}
    \bm M_i^{t+1} = k_i\left[\bm M_i^t + \frac{\bm S_\mathrm{p}^t + \bm S_\mathrm{n}^t}{2}\right]_0^1
\end{equation}  
where $\bm M_i^t$ indicates the $i$-th channel of the memory of events at time $t$, $k_i \in (0, 1)$ is the forgetting factor of the $i$-th channel, $\bm S_\mathrm{p}^t$ and $\bm S_\mathrm{n}^t$ are the positive and negative time surfaces at time $t$ and operator $\left[\cdot\right]_0^1$ saturates the argument between 0 and 1. 
In this work we use three channels with forgetting constants $k_1=0.8$, $k_2=0.6$ and $k_3=0.4$, so the input of the DNN estimator is a tensor of shape $[3, H, W]$. 
        
The enriched events representation allows to use classical CNNs architectures that are already optimized for edge applications. In this work, we choose MobileNet-V3L, a widely known lightweight DNN already used in production-grade applications. This model is tuned for edge CPU-based devices thanks to a combination of hardware-aware network architecture search (NAS) and pioneering architectural enhancements \cite{howard2019searching}.
        
\paragraph{Implementation Details.}

The proposed approach moves the time dependency of our model entirely to the input pipeline, meaning that it is possible to apply to this task any non-recurrent (time-independent) vision model. In particular, we utilized MobileNet-V3L \cite{howard2019searching} pretrained on ImageNet-1K, and we fed the model with information at different time resolutions by combining multiple memory of events, with different forgetting factors, as input channels. We report in Table~\ref{tab:complexity} details about the time required to create the memory of events used as input of the model and the inference time of the network.

\begin{figure}[ht]
    \centering
    \includegraphics[width=3.3in]{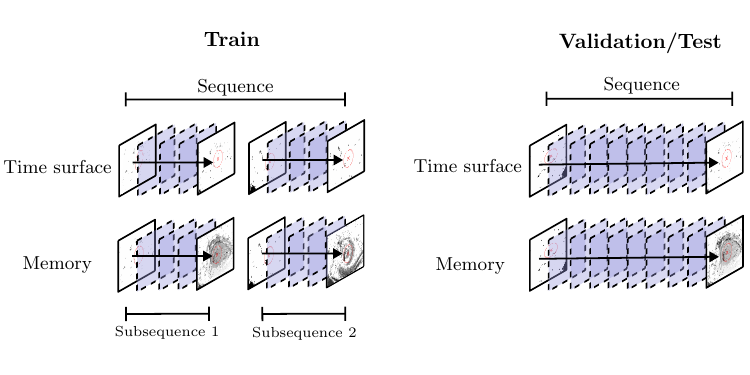}
    \caption{For training, original sequences of time surfaces are divided in subsequences which are then fed to the DNN in a random order. For validation/test data, the original sequence is not subdivided to emulate the behavior of the final system.}
    \label{fig:train}
\end{figure}

For training, we used only the 2024 Event Eye Tracking Challenge (EET) dataset. Each event stream is divided into small chunks containing events in a time range $\Delta t =50ms$. From each chunk, a time surface is created. In Figure~\ref{fig:train} the term ``Sequence'' indicates the collection of all the time surfaces extracted from a recording. 
During training, these time surfaces are organized in subsequences, as constructing meaningful input memories of events requires consecutive time surfaces. The subsequences are then fed to the DNN in a random order. For what concerns the validation and the test set, each sequence is not subdivided into smaller subsequences to emulate the behavior of the system. 

The network underwent training for 200 epochs using the Adam optimizer~\cite{kingmaAdamMethodStochastic2017} and an initial learning rate of \num{2.8e-4}. The training process lasts almost 2 hours on a V100 GPU.
    
\begin{table}[t]
    \centering
    \caption{Computational complexity of the proposed method}
    \label{tab:complexity}
    \resizebox{0.5\textwidth}{!}{
    \begin{tabular}{ccc}
    \toprule
        \textbf{} & \textbf{FLOPS} & \textbf{Inference time}\\
        \midrule
        Memory creation & \SI{24}{\kilo\relax} & \SI{0.1}{\milli\second} (CPU -- Intel Xeon Gold)\\
        MobileNet-V3L & \SI{0.23}{\giga\relax} & \SI{4.9}{\milli\second} (GPU -- Nvidia A100)\\
        \bottomrule
    \end{tabular}
    }
\end{table}
    
The best-performing model, in terms of validation accuracy (p10), was saved during training. In addition, to ensure robustness and avoid over-fitting, we largely used data augmentation on the training set.

\paragraph{Results.}
We evaluate the proposed system based on memory channels on the full test set. The results obtained are the following: 99.53\% of p15 accuracy, 99.05\% of p10 accuracy, 89.36\% of p5 accuracy and a mean euclidean distance of 3.2. 

Moreover, we also evaluate the effectiveness of our pre-processing methodology against the use of simple time surfaces as inputs. When using time surfaces only, the input of the estimator is composed of three channels, which are $\bm S_\mathrm{p}$, $\bm S_\mathrm{n}$ and their average.
Table~\ref{tab:deltat_res} compares the performance of MobileNet-V3L on the full test set of the EET dataset, measured by means of the p10 accuracy and the mean Euclidean distance with $\Delta t = \SI{50}{\milli\second}$. The results consistently demonstrate a significant improvement when using memory channels over simple time surfaces.

\begin{table}[t]
  \caption{P10 accuracy and Mean Eucledian distance of the eye-tracking system on the full EET test set, both with the use of time surfaces only or incorporating the input pipeline based on three memory channels with $\Delta t=\SI{50}{\milli\second}$.
  }
  \label{tab:deltat_res}
  \setlength{\tabcolsep}{4pt} 
  \resizebox{0.5\textwidth}{!}{
  \begin{tabular}{ccccc} 
    \toprule
    \multicolumn{1}{c}{} & \multicolumn{2}{c}{\textbf{p10 Accuracy}} & \multicolumn{2}{c}{\textbf{Mean Eucledian distance}}                   \\
    \cmidrule(r){2-3} \cmidrule(l){4-5}
                 & \begin{tabular}{c}TS only\end{tabular} & \textbf{Ours} & \begin{tabular}{c}TS only\end{tabular} & \textbf{Ours} \\
    \midrule
    MobileNet-V3L     &      94.9\%             & \textbf{99.1\%} & 3.7            & \textbf{3.2}    \\
    
    \bottomrule
    
  \end{tabular}
  }
\end{table}


As far as we know, the input pipeline we propose based on memory channels is novel and the entire proposed solution is production-ready since it uses a DNN that is already optimized for edge devices, while the input pipeline is implementable in a few lines of code.

\begin{figure*}[t]
    \centering
    \includegraphics[trim={0 1.5cm 0 0},clip, width=\textwidth]{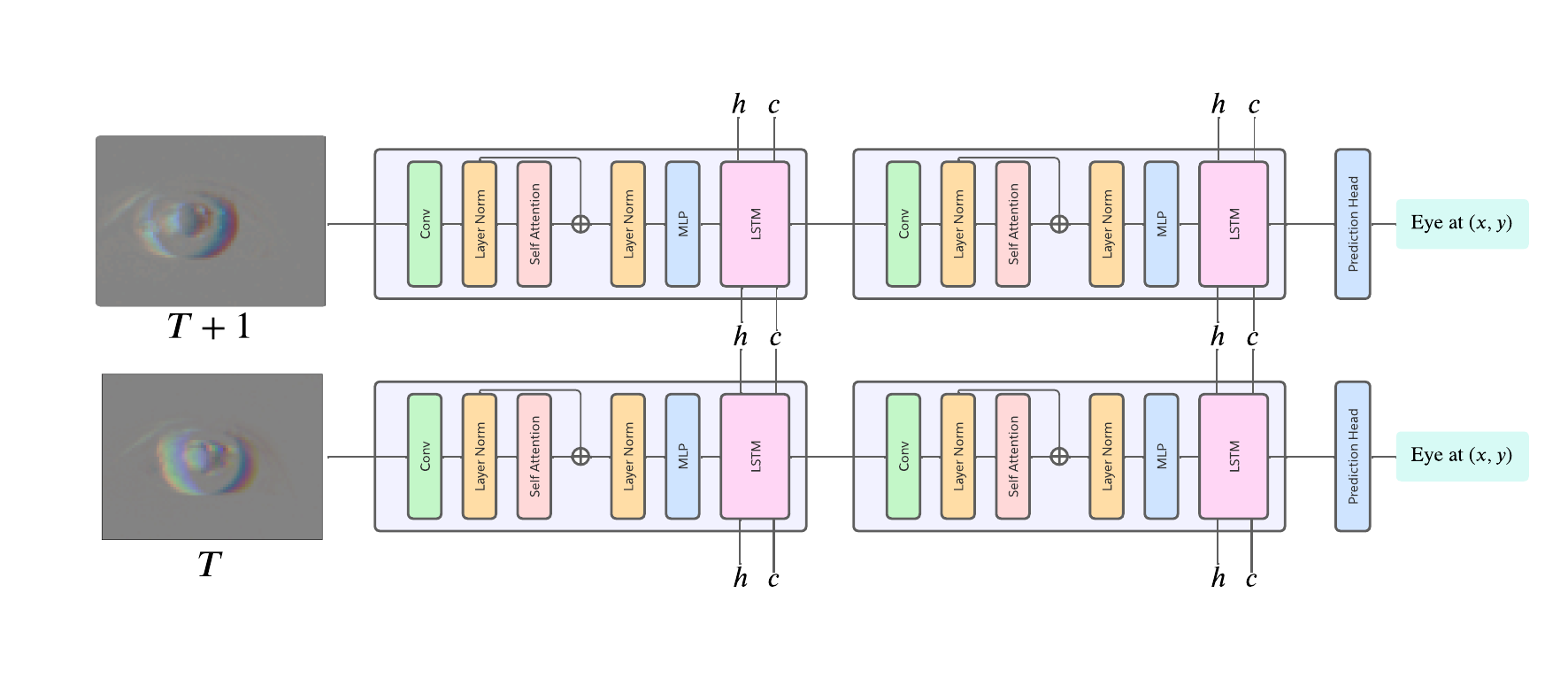}    
    \caption{Model architecture diagram proposed by Team ERVT, showing the two stages of the model and over two timesteps.}
    \label{fig:my_diagram}
\end{figure*}

\subsection{Team: ERVT}


\begin{center}
\noindent\emph{Mircea Lică, David Dinucu-Jianu, Cătălin Grîu}

\noindent\emph{Delft University of Technology}

\noindent{\emph{Contact: \url{M.T.Lica@student.tudelft.nl}}}   
\end{center}


\paragraph{Description.} 
We propose a lightweight model, Efficient Recurrent Vision Transformer (ERVT), inspired by prior work in event-based object detection \citep{gehrig2023recurrent}, tailored for real-time eye tracking applications. To accurately predict the position of the pupil, we designed a hierarchical model based on multiple stages working at different feature resolutions. Each stage employs a spatial feature extraction module and a spatio-temporal module. Compared to the Recurrent Vision Transformer \cite{gehrig2023recurrent}, ERVT does not completely separate spatial and temporal feature extraction, employing a ConvLSTM to capture spatio-temporal relations between consecutive timestamps. Moreover, we replace the block-grid attention introduced in \citep{tu2022maxvit} with pixel-wise multi-head self-attention to better capture global information. ERVT achieves a P10 accuracy of $97.6\%$ on the private competition dataset with only 150K parameters, $37$M MACs, $0.157$ GFLOPs and an inference time of $1.06$ms on a RTX 3060 Laptop GPU.

\paragraph{Implementation Details.}
The Efficient Recurrent Vision Transformer (ERVT) is an adaptation of the Recurrent Vision Transformer (RVT) \cite{gehrig2023recurrent} with an emphasis on efficiency in terms of inference time and size, such that it could ultimately be used in real-time applications. The model is composed of two stages that operate at different spatial resolutions and followed by a prediction head which produces the $(x, y)$ coordinates corresponding to the pupil. Each stage of the network consists of a strided convolution, a spatial feature extraction module and a ConvLSTM \cite{shi2015convolutional}, as seen in Fig~\ref{fig:my_diagram}. The stage starts with a strided convolution in order to exploit the inductive biases given by the spatial structure of the image while downsampling the feature maps for more efficient computation. The spatial feature extraction module takes inspiration from \cite{tu2022maxvit} and is designed to capture global relationships between pixels in the corresponding feature maps using self-attention. Contrary to the original Recurrent Vision Transformer, we find that the model performs significantly better without completely separating spatial and temporal feature extraction. Thus, each stage ends with a ConvLSTM that aggregates the information extracted at the current timestamp with the information from the previous timestamp. \\ 

The model is trained from scratch on the dataset provided by the competition. Similarly to the baseline provided by the organizers, we pre-process the event stream into a voxel grid representation with $T = 3$
discretization steps of time to make it compatible with the convolutions required for ERVT. During training, we split the dataset in sequences of 30 frames (voxel grids) corresponding to $50$ms of accumulated events. To alleviate the problem of exploding/vanishing gradients specific to RNNs, we use a Truncated Backpropagation Through Time (TBPTT) optimization technique. At inference time, we process frames in a sequential manner, keeping track of the hidden $h$ and cell $c$ states of the ConvLSTM for the next timestamp. \\

We apply both temporal and spatial sub-sampling with a factor of $\alpha = 0.2$ and $\beta = 0.125$ for the temporal and spatial dimension, respectively. To limit overfitting on the dataset, we augment the data with 2 different techniques: random horizontal flip and random noise. The horizontal flip has a probability of $0.5$ while the noise is applied pixel-wise based on a normal distribution $\mathcal{N}(\mu, \sigma^2)$ where the mean and standard deviation are applied per input channel. Our method does not use the eye open/closed information additionaly provided in the dataset. \\

We use PyTorch for training and inference of ERVT and Tonic library to process the event dataset provided by the competition. The model is trained to minimize the weighted RMSE loss between the predicted $(\hat{x}, \hat{y})$ and the ground truth $(x, y)$ using Adam \cite{kingma2014adam} as the optimizer with an exponential learning rate scheduler with $\gamma = 0.98$. \\

The training procedure consists of $150$ epochs over the competition dataset with data augmentation included. The initial learning rate is set to $0.001$ and we use a batch size of $1$ to avoid overfitting on the train set. As mentioned before, truncated backpropagation through time is used as an optimization strategies, with the goal of removing the effects of exploding/vanishing gradients caused by the recurrent component. Thus, we split the sequence of $30$ frames in two equal parts and stop the gradients when passing from one part to the next, as the standard TBPTT is implemented. Using these settings, the training time for ERVT is $5.4$ hours, Lastly, we use a dropout probability of $0.6$. All the experiments are performed on a laptop with Nvidia RTX 3060 6 GB.




\subsection{Team: EFFICIENT}
\label{sec:team:efficient}

\begin{center}
\noindent\emph{Xiaopeng Lin, Hongwei Ren, Bojun Cheng}

\noindent\emph{The Hong Kong University of Science and Technology (Guangzhou)}

\noindent{\emph{Contact: \url{bocheng@hkust-gz.edu.cn}}}   
\end{center}

\begin{figure*}[tb]
    \centering
    \includegraphics[width=\textwidth]{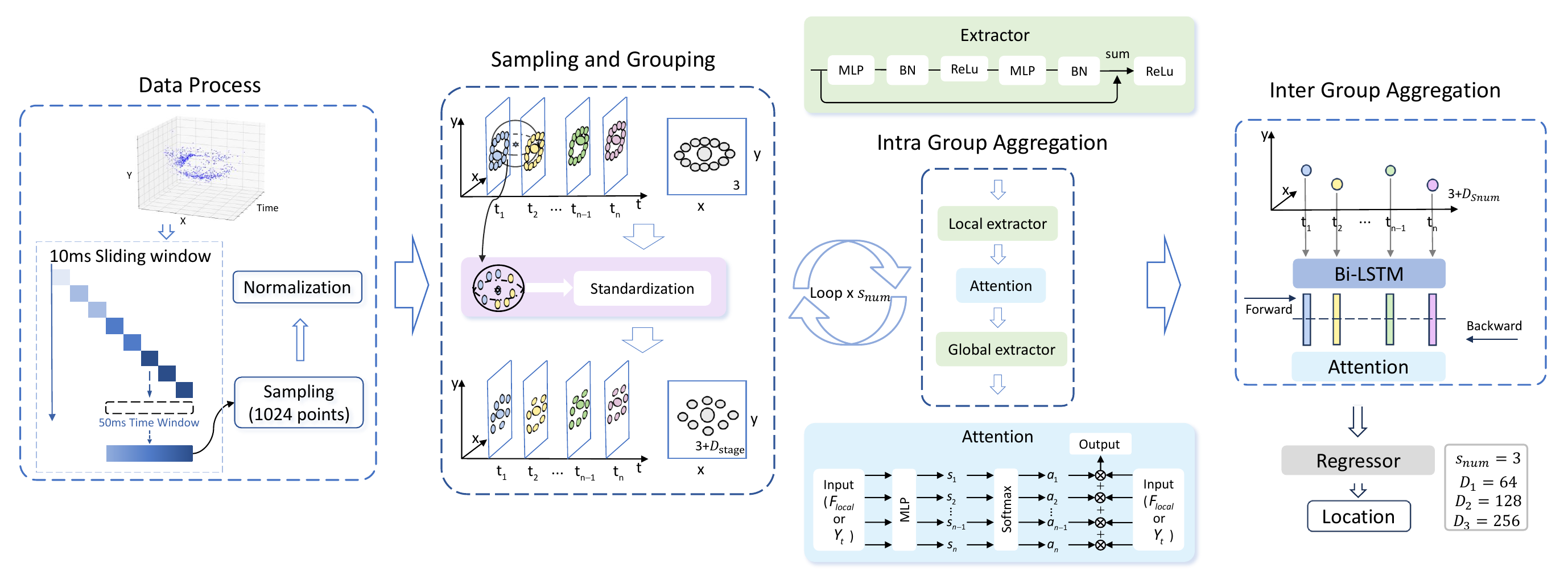}
    \caption{Efficient Point-based Eye Tracking Network.  The input Event Cloud is directly processed using a sliding window, sampling, and normalization, eliminating the necessity for any format conversion.  Sequentially, the input passes through $S_{num}$ hierarchy structures for spatial feature abstraction and extraction.  Furthermore, the input is then passed through a bidirectional LSTM to extract temporal features, culminating in a regressor responsible for eye tracking.}
    \label{fig:NETWORK_efficient}
\end{figure*}
\paragraph{Description.} 
       The EFFICIENT team utilizes the efficient Point-based Network named PEPNet \cite{ren2024simple} for Eye Tracking with Event Cameras. The raw point cloud is directly leveraged as the network input to fully utilize the high-temporal resolution and inherent sparsity of events. An Intra Group Aggregation Module is designed to extract the local spatial and temporal features. The global information is aggregated through Inter Group Aggregation Module, which consists of the Bi-directional Long Short-Term Memory. The Point-based Network can achieve excellent performance at high inference speed with a lightweight design principle, only 0.64 Million parameters and the inference time is 6.7 ms/sample. Additionally, the parameter and floating-point operations (FLOPs) in our model remain constant regardless of the input camera's resolution, in contrast to frame-based methods.
       
       The network architecture is shown in Fig. \ref{fig:NETWORK_efficient}, encompassing Data Processing, Sampling and Grouping, Intra-Group aggregation, Inter-Group aggregation, and Regression. The initial phase leverages the sparsity and high temporal resolution of event data, processing the event stream into non-overlapping 10 ms samples, each associated with a 100 Hz label. To address challenges associated with data sparsity, the methodology includes expanding the sample window size to 50 ms, maintaining the original labeling frequency to align with evaluation metrics. The Sampling and Grouping phase employs advanced techniques such as Farthest Point Sampling (FPS) and K-Nearest Neighbors (KNN) to capture essential spatial and temporal information efficiently \cite{qi2017pointnet++}.
       
       To enhance feature extraction and integration, the methodology includes two novel aggregation modules. The Intra Group Aggregation module leverages an extractor coupled with an attention mechanism to distill and consolidate local spatial-temporal information. Concurrently, the Inter Group Aggregation module employs a Bi-directional LSTM network, synergized with an attention mechanism, to elucidate temporal relationships across different event groups. This bifurcated aggregation strategy enables a comprehensive synthesis of local and global features, facilitating precise pupil position regression within a lightweight model architecture.
       
       The Efficient Point-based Eye Tracking Method exhibits capabilities for high-frequency tracking operations. Leveraging a lightweight asynchronous architecture, this methodology demonstrates remarkable adaptability. It enables precision tracking across a diverse range of frequency bands, with operational effectiveness maintained up to frequencies as high as 1 kHZ. This adaptability, coupled with its high-frequency tracking capabilities, positions the method as a robust solution for applications requiring precise eye movement tracking under varying operational frequencies.
\paragraph{Implementation Details.}

Our server leverages the PyTorch deep learning framework and selects the AdamW optimizer with an initial learning rate set to $1\cdot e^{-3}$, which is reduced at the 100th and 120th epochs, accompanied by a weight decay parameter of $1\cdot e^{-4}$. This configuration is meticulously chosen to enhance the model's convergence and performance through adaptive learning rate adjustments. Training is conducted on an NVIDIA GeForce RTX 4090 GPU with 24GB of memory, enabling a batch size of 256. The model size is contingent upon the dimensionality of MLPs at each stage. The MLPs’ dimensions for the standard structure are [64, 128, 256]. Moreover, the Bi-LSTM hidden layer dimension is 128.

The validation and testing protocol is meticulously designed to assess the solution's accuracy and robustness comprehensively. Testing involves processing motion data into non-overlapping 50 ms samples for 20 Hz tracking, with a pre-processing step of random sampling 1024 event points per sample. Initial tests without post-processing achieves 93\% P10 accuracy, which is further enhanced through a novel confidence score mechanism, boosting the tracking accuracy to over 98\% P10 accuracy. This mechanism accounts for the number of event points per sample, improving prediction reliability significantly.


\paragraph{Results.}
The Efficient Point-based Eye Tracking Method showcases impressive performance across various metrics, achieving P5 accuracy at 80.67\%, P10 at 97.95\%, and P15 at 99.74\%, with a mean Euclidean distance of 3.51 and a mean Manhattan distance of 4.43. The innovation of this research lies in its effective utilization of point clouds within the event-based eye tracking paradigm, exploiting the high temporal resolution of event data to achieve high tracking frequency and accuracy on a streamlined model. This innovative approach enables high-frequency, high-accuracy tracking on a lightweight model, marking a significant contribution to the field of eye tracking technology.
\subsection{Team: GTechVision}


\begin{center}
\noindent\emph{Xinan Zhang, Valentin Vial, Anthony Yezzi, James Tsai}

\noindent\emph{
Georgia Institute of Technology
}

\noindent{\emph{Contact: \url{xzhang979@gatech.edu}}}
\end{center}


\paragraph{Description.} 
Our methodology can be divided in three steps :
        
\begin{itemize}
    \item With the real-time edge applications taken into account, we adopted and adjusted three light-weighted models for this task, including the baseline (CNN+GRU), convolutional LSTM network, and spiking LSTM model based on spiking neural network.
    
    \item A comprehensive hyper-parameter search on these Model has been performed, resulting in the identification of hyper-parameter impacting performances and further selection of the best combination of the hyper-parameter.
    
    \item The four-layer convolutional LSTM network achieved the best performance with the selected hyper-parameters, with a moderately high p10 accuracy of 92.63 percent and fast inference speed less than 1ms per event on a RTX 3090 Ti GPU.
\end{itemize}

In our methodology, the principle of grid search plays a pivotal role, unfolding in two distinctive dimensions. First, we confronted the constraints set by the realities of edge computation in real-world decentralized applications, leading us to limit our model selection to those with a size of generally less than twenty million and the final submitted model has merely 0.41 million learnable parameters. Among these, the baseline, convolutional LSTM \cite{shi2015convolutional, 3et}, and spiking LSTM \cite{SpikingLSTM} stands out as notably effective and lightweight, drawing from insights gleaned from prior literature on analogous tasks.

Subsequently, to optimize the performance of these chosen models to their fullest potential, we conducted a meticulous grid search across a spectrum of hyperparameters, local structures, loss functions, etc. This systematic exploration aimed to fine-tune the models, ensuring to boost their performance in event based eye tracking. The baseline, convolutional LSTM, spiking LSTM will be discussed below with different emphasis. 

\begin{figure}[h]
  \centering
  \includegraphics[width=0.5\textwidth]{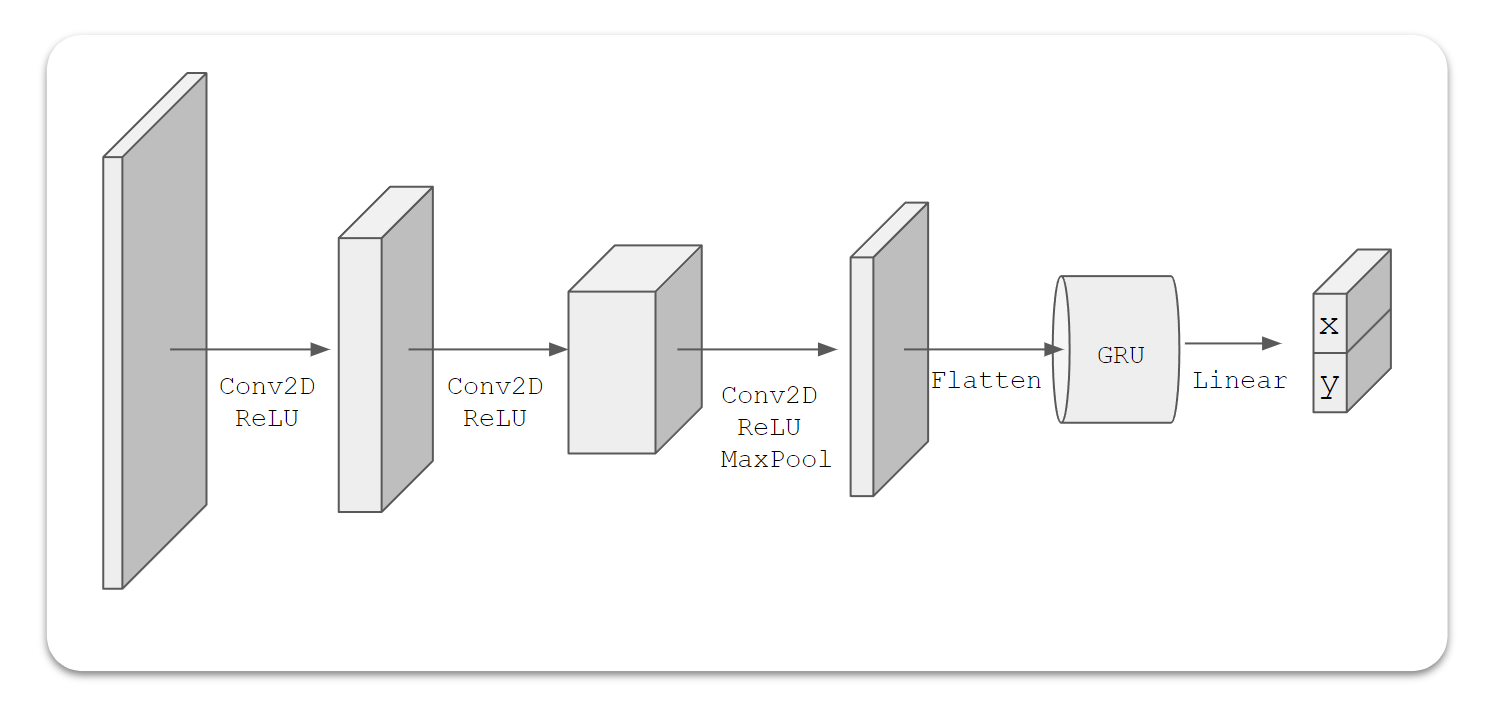}
  \caption{Baseline Architecture}
  \label{fig:baseline_gt}
\end{figure}

The baseline, as shown in Fig.\ref{fig:baseline_gt} consists of three layers of 2D convolution as the backbone and one layer of GRU to extract the semantic information and further the temporal dependency between adjacent timestamps. Finally, one fully connected layer regress the pupil center coordinates. As mentioned above, a set of grid search has been performed and it shows that 6 time dimension channels in the voxel representation can yield best p10 performance for the baseline with L1 loss.

\begin{figure}[h]
  \centering
  \includegraphics[width=0.5\textwidth]{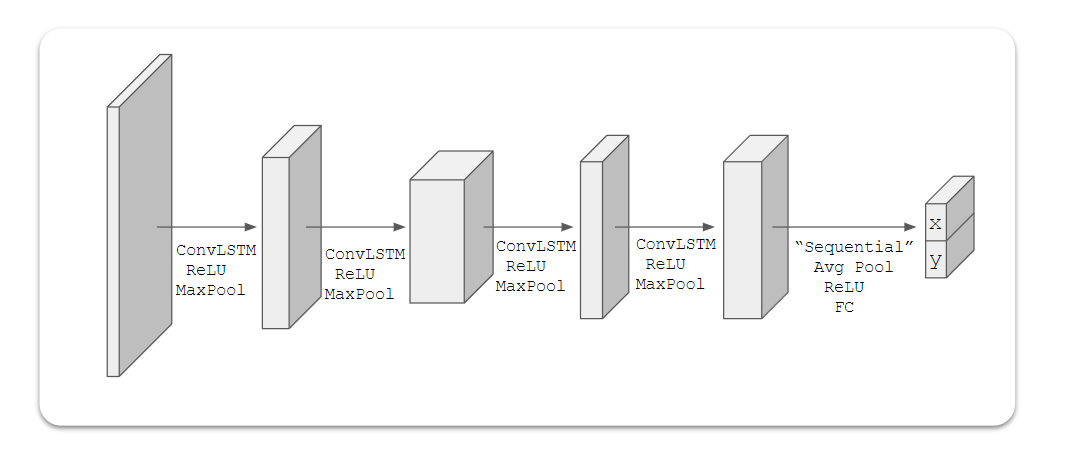}
  \caption{Convolutional LSTM Architecture}
  \label{fig:convLSTM_gt}
\end{figure}

The second model we reported here is convolutional LSTM model in Fig.\ref{fig:convLSTM_gt}, with convolutional LSTM layer as its fundamental building block, which takes in the feature maps at the current timestamp from the previous layer (voxel representation if it is the first layer) and the previous hidden feature (a list of zero tensors if it is the first layer), utilizes 2D convolution layers to extract high-level feature maps given to the LSTM to generate new hidden features. The newly generated hidden features will be used for regressing the coordinates as well as input to the next convolutional LSTM cell sequentially. We experimented with different structures for the feature map (2D convolution and 3D convolution) and temporal dependency extraction (LSTM, Bi-LSTM), different number of convolutional LSTM layers (2 ,3, 4) and loss functions (MSE, Weighted MSE, L1) and finally found that 4 layers with each consisting of 2D convolution and uni-directional LSTM gives the best p10 performance with 3 time channels in the voxel representation and Weighted MSE loss.

\begin{figure}[h]
  \centering
  \includegraphics[width=0.5\textwidth]{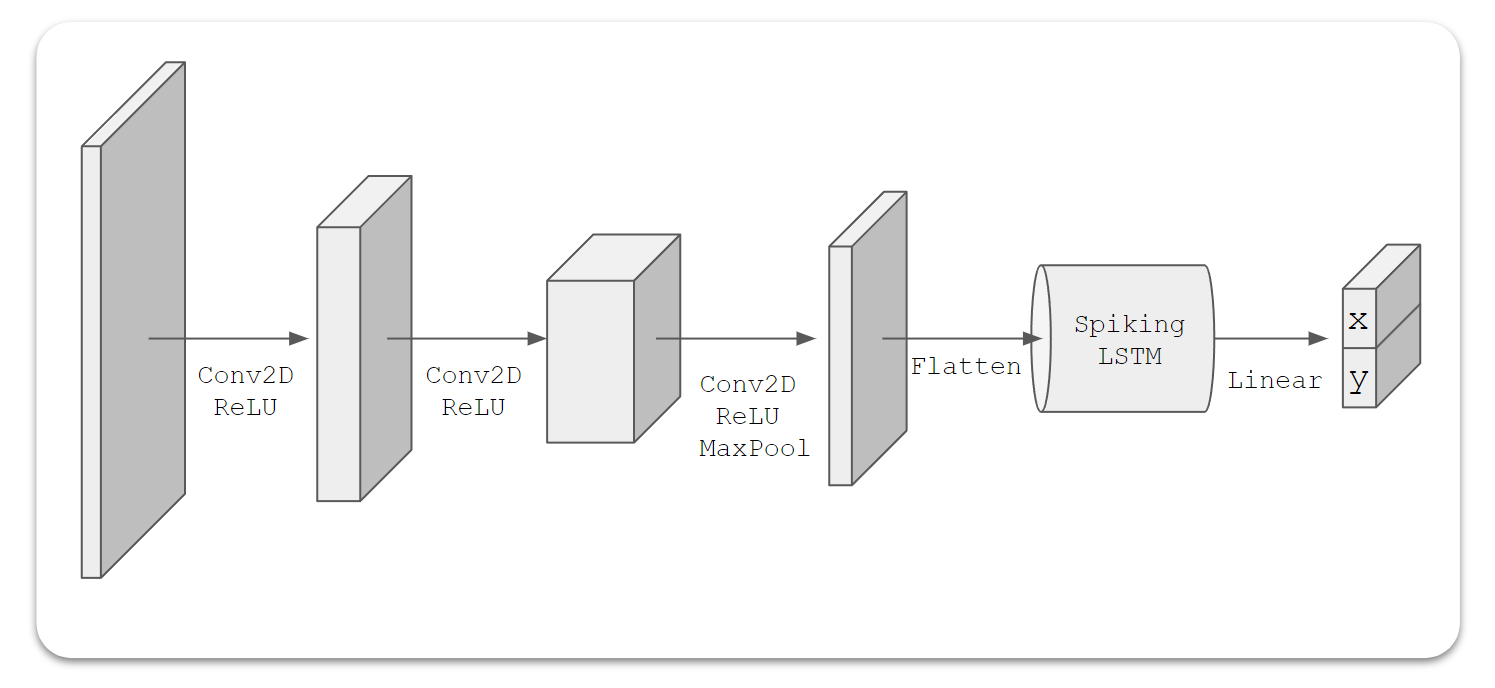}
  \caption{Spiking LSTM Architecture}
  \label{fig:snn_gt}
\end{figure}

We also explored the spiking LSTM model as shown in Fig.\ref{fig:snn_gt}, which shares a similar structure with the baseline but uses a spiking LSTM instead of GRU to extract the temporal dependency. In Spiking LSTM, a set of spiking activation functions are applied to neurons at forget gate, input gate and another assisting layer on input. Membrane potential of the neurons are inputs to the spiking activation functions which eventually output a spike or null in the order of a time series. Further, the activation functions in the backbone were also experimented with either regular ReLU or Leaky Integrate-and-Fire (LIF) Neurons. The final results show that integration of spiking neurons and a spiking LSTM architecture by 5\%, reaching 87.68\% P10 accuracy with 6 time channels in the voxel representation and L1 loss.

\paragraph{Implementation Details.}
We adhered to the data preparation pipeline outlined in the official demo, which involved spatial and temporal sampling, division of sub-recordings, aggregation of sparse events over consecutive frames to construct a voxel representation within one sub-recording. This approach allowed us to condense both spatial and temporal information within the prepared data. The voxel representations in a sub-recording are inputted to the networks sequentially as a time series, which are expected to regress the coordinates while considering the temporal dependency between adjacent voxel representations. Besides, the annotated data was divided into training and validation sets subsequently. Notably, we refrained from incorporating any extra data beyond the original one but decreased the default train stride from 15 to 5, which means the next sub-recording will start 250us after the current one instead of 750us, in order to generate more data for training from the raw recordings.
For the training process, we employed a single RTX 3090 Ti GPU. Utilizing the Adam optimizer, we set the learning rate at 0.001, and the training proceeded through a total of 200 epochs. This setting also aligned with the demo.

\paragraph{Results.}

The trio of models we selected provides a natural framework for discerning the difference between various RNN structures. Within this set, the baseline model utilizes the GRU instantiation, while its counterparts leverage the convolutional LSTM and spiking LSTM architectures respectively. The best P10 performance of the three models are reported in Table \ref{tab:eye_tracking_solution_gt}.

\begin{table}

\resizebox{0.5\textwidth}{!}{
\begin{tabular}{lrlllll}
\toprule
 Method & Optimizer & Learning Rate & Epochs & Dataset & P10 Accuracy \\
\midrule
4-Layer Convolutionnal LSTM & Adam & 0.001 & 200 & Challenge dataset & 92.63\%\\
Spiking LSTM & Adam & 0.001  & 200 & Challenge dataset & 87.68\%  \\
Baseline & Adam & 0.001  & 200 & Challenge dataset  & 83.99\% \\
\bottomrule
\end{tabular}
}
\caption{P10 Performance of The Three Models by Team GTechVision.}
\label{tab:eye_tracking_solution_gt}
\end{table}

Notably, the convolutional LSTM model emerged as the top performer among the three, with the fewest trainable parameters, which underscores the efficacy of employing convolutional LSTM within a constrained model size. The strength of the convolutional LSTM architecture lies in its ability to retain more spatial information within the LSTM cell  with 2D convolution so that the feature maps are flattened at later stages than other models, which could be beneficial to 2D coordinate regression and future timestamps. Furthermore, our findings also shed light on the promise of the spiking LSTM model in this task. While not superior to the convolutional LSTM model, it outperformed the baseline by 3.69\% in p-10 accuracy.

{\small
\bibliographystyle{ieeenat_fullname}
\bibliography{shared_bib,bibs/bigBrains,bibs/Efficient,bibs/ERVT,bibs/FreeEvs,bibs/gosparse,bibs/GTechVision,bibs/trashcoders,bibs/USTCEventGroup}
}

\end{document}